%% file: main.tex
\documentclass[letterpaper, 10 pt, conference]{ieeeconf}  

\IEEEoverridecommandlockouts                              

\usepackage{amsmath,amsfonts}
\usepackage{algorithmic}
\usepackage{algorithm}
\usepackage{array}
\usepackage{textcomp}
\usepackage{stfloats}
\usepackage{url}
\usepackage{verbatim}
\usepackage{graphicx}
\usepackage[nocompress]{cite}
\usepackage{adjustbox}
\usepackage{booktabs}
\usepackage{multirow} 
\usepackage{tabularx}
\usepackage{makecell}
\usepackage{pifont}
\usepackage{xcolor}
\usepackage{float}
\usepackage{cuted}     
\usepackage[hidelinks]{hyperref}
\usepackage[nameinlink,noabbrev]{cleveref} 

\let\labelindent\relax
\usepackage{enumitem}

\newlist{compactitem}{itemize}{1}
\setlist[compactitem]{
    label=\textbullet,
    leftmargin=*,
    topsep=2pt,
    itemsep=1pt,
    parsep=0pt,
    partopsep=0pt
}

\crefname{figure}{Fig.}{Figs.}
\Crefname{figure}{Fig.}{Figs.}
\crefname{table}{Table}{Tables}
\Crefname{table}{Table}{Tables}
\crefname{section}{Sec.}{Secs.}
\Crefname{section}{Sec.}{Secs.}
\crefname{subsection}{Sec.}{Secs.}
\Crefname{subsection}{Sec.}{Secs.}
\crefname{subsubsection}{Sec.}{Secs.}
\Crefname{subsubsection}{Sec.}{Secs.}
\crefname{equation}{Eq.}{Eqs.}
\Crefname{equation}{Eq.}{Eqs.}

\newcommand{\method}{\textsc{DexOAK}}

\newcommand{\csdone}[1]{\textcolor{orange}{CS done: #1}}

\renewcommand{\csdone}[1]{}    

\newcommand{\Real}{\mathbb{R}}

\newcommand{\SEThree}{\mathrm{SE}(3)}
\newcommand{\SOthree}{\mathrm{SO}(3)}

\def\BibTeX{{\rm B\kern-.05em{\sc i\kern-.025em b}\kern-.08em
    T\kern-.1667em\lower.7ex\hbox{E}\kern-.125emX}}

\author{Federica Spinola$^{1}$ and Robert Katzschmann$^{2}$ and Cordelia Schmid$^{1}$ \\
\url{https://fspinola.github.io/dexoak/}
\thanks{$^{1}$ Inria and the Departement d’Informatique of Ecole Normale Supérieure, PSL University, France
        {\tt\small federica.spinola@inria.fr, cordelia.schmid@inria.fr}}%
\thanks{$^{2}$Soft Robotics Lab, ETH Zürich, Switzerland
        {\tt\small rkk@ethz.ch}}%
}

\title{\LARGE \bf
\method{}: Object-Aware Kinematic Retargeting
}

\begin{document}

\maketitle
\thispagestyle{empty}
\pagestyle{empty}

\input{sections/00_abstract.tex}


\input{sections/01_introduction.tex}

\input{sections/02_related_work.tex}

\input{sections/04_method.tex}

\input{sections/05_experiments.tex}

\input{sections/08_conclusion.tex}





\section*{Acknowledgments}

This work was granted access to HPC resources of IDRIS
under the allocation AD011017024 made by GENCI. It was
funded in part by the French government under management
of Agence Nationale de la Recherche as part of the “France
2030” program, reference ANR-23-IACL-0008 (PR[AI]RIEPSAI project) and the ANR project VideoPredict (ANR-21-FAI1-0002-01). Cordelia Schmid would like to acknowledge the support by the K\"orber European Science Prize. The authors thank Peter Kulits, Sara Pieri and Seok Joon Kim for their valuable contributions to this project. Finally, the authors would like to acknowledge that Claude (Anthropic, Opus 5.5 and Fable 5) was used to assist in writing portions of the code and to proofread the text for grammar and clarity. The tool was not used to generate results or conclusions. The authors reviewed all AI-assisted output and take full responsibility for the final content.

\bstctlcite{IEEEexample:BSTcontrol}
\bibliographystyle{IEEEtran}
\bibliography{IEEEtranBST/IEEEabrv,IEEEtranBST/references}

\input{sections/supplementary}

\end{document}

%% file: sections/00_abstract.tex
\begin{abstract}

We introduce \method{}, a learning-based object-aware kinematic retargeting framework for generating contact-consistent robot hand references from human motion-capture data. \method{} first minimizes an interaction-graph Laplacian objective over a joint hand--object graph, so the robot reproduces the demonstrated object-relative interaction geometry rather than the human hand pose alone.
These per-frame solutions form seed retargeted references, from which a feed-forward network conditioned on human keypoints and object point clouds learns the retargeting function. Object-aware fine-tuning then refines the network under contact, non-penetration, and self-collision losses that position the fingers on the demonstrated contacts and reduce artifacts destabilizing physics-based refinement.
Once trained, the network transfers to unseen data by inference alone, without per-frame optimization.

\method{} references improve dynamic retargeting over the state-of-the-art across bimanual tasks with rigid and articulated objects.
On the Inspire hand, success rises from 39.5\% to 48.4\% on OakInk-v2, and ADD-AUC rises from 62.7\% to 77.9\% on ARCTIC.
Ablations trace the gains to object-aware kinematic retargeting and the amortized network refined with contact losses.
Trained on OakInk-v2, it transfers to ARCTIC zero-shot and nears in-domain performance after fine-tuning. On hardware, 46/56 open-loop replays succeed across five tasks, enabling dexterous robot data generation from human data.

\end{abstract}

%% file: sections/01_introduction.tex
\section{Introduction}
\label{sec:intro}

Progress in robot manipulation increasingly comes from learning at scale, yet the resulting policies~\cite{openvla,pi07} remain bounded by scarce robot interaction data. It cannot be scraped from the web like the image and text corpora behind vision and language models, and teleoperation, the most reliable alternative, is slow and costly. The gap is widest for dexterous manipulation, where multi-fingered hands are hard to teleoperate~\cite{dexcap}, leaving contact-rich interaction especially scarce. Abundant clips of people handling objects, increasingly annotated with hand and object pose~\cite{dexycb,arctic,oakinkv2}, are a natural source of the dexterous interaction data that robots lack.
Such data supports robot learning through representation learning~\cite{r3m}, policy pretraining~\cite{egovla}, imitation learning~\cite{mimicplay}, or retargeting~\cite{spider}. We focus on retargeting because it directly converts human demonstrations into dynamically feasible robot trajectories~\cite{maniptrans, dexmachina}, bypassing teleoperation entirely.

A human-to-robot retargeting pipeline has two stages. Kinematic retargeting sets \emph{where} the robot hand goes relative to the environment, and dynamic retargeting sets \emph{how} it actuates to interact with the object.
Kinematic retargeting most often matches the robot and human vectors between hand keypoints~\cite{anyteleop,dexpilot}, fitting hand shape alone and ignoring the object. Yet human and robot hands differ in degrees of freedom (DoF) and embodiment geometry, so shape similarity does not guarantee task-equivalent interaction.
Instead, recent work~\cite{omniretarget} preserves robot--environment relations with per-frame interaction-graph optimization, but targets whole-body loco-manipulation rather than dexterous finger-level contact.

\begin{figure}[t]
    \centering
    \includegraphics[width=\linewidth]{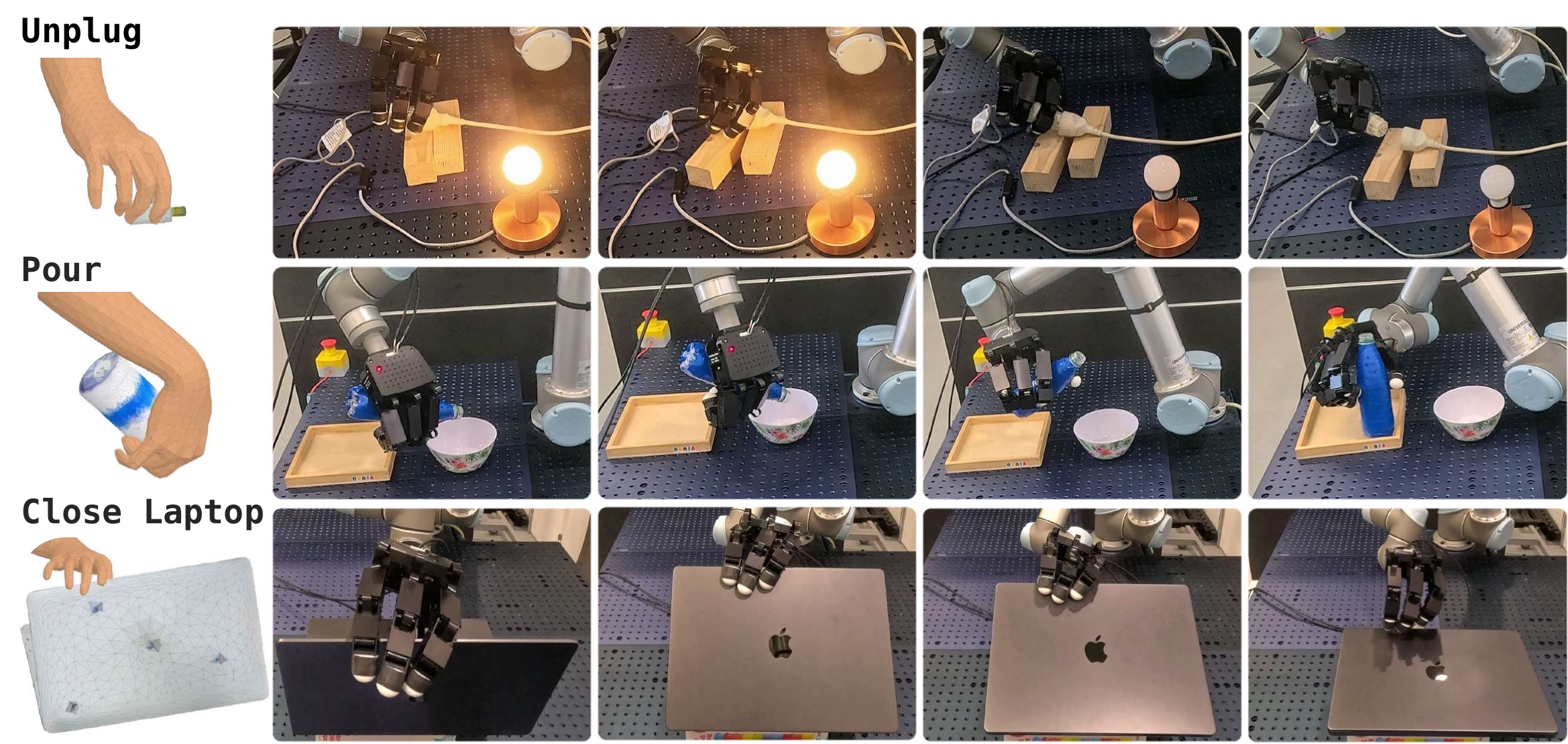}
    \caption{\textbf{\method{} enables dexterous real-robot data generation from human motion-capture.}
Human demonstrations of \emph{Unplug}, \emph{Pour}, and \emph{Close Laptop} (held out from retargeter training) are successfully replayed open-loop on a UR5 + right Allegro using \method{} references refined by RL.
}
    \label{fig:teaser}
\end{figure}

We argue that kinematic retargeting should preserve task function rather than hand-pose fidelity. What matters is the human--object interaction: object-relative fingertip placement, grasp approach, and object motion. Prior pipelines introduce object and contact information only during physics-based refinement~\cite{dexmachina,maniptrans,spider}, and never refine the kinematic reference itself under contact objectives.
We therefore propose \method{}, which learns object-aware kinematic references
before dynamic refinement.
Every reference still requires physics refinement, but its quality determines how much error the dynamic stage must correct. We hypothesize that object-aware references let it focus on dynamics rather than inferring contact locations, improving performance.

\method{} is a two-stage pipeline that turns human hand--object motion-capture demonstrations into kinematic robot trajectories. An optimization-based retargeter builds paired human--object and robot--object
interaction graphs~\cite{omniretarget} and solves for robot wrist pose and joint angles matching their Laplacian coordinates.
Solving these per-frame nonlinear programs sequentially yields object-aware kinematic references, our
\emph{seed retargeted references}. However, this optimization is sensitive to initialization~\cite{numericaloptim} as contact and collision terms would make
the objective more non-convex. 
We therefore train a differentiable network, sharing structure across tasks, to reproduce the seed references with one forward pass per frame instead of
per-frame optimization. This supervised amortization initializes \emph{object-aware fine-tuning}, whose surface-contact,
object-part, and contact-direction losses improve robot finger placement beyond
the optimizer outputs. Importantly, the fine-tuned
network transfers to unseen data without a single per-frame solve. 
\method{}'s kinematic references act as the base action for a residual
reinforcement learning (RL) stage~\cite{johannink2018residual, maniptrans} that corrects the remaining dynamics gap. Successful simulation rollouts are then replayed open-loop on hardware, demonstrating physical executability and yielding real-robot trajectory data from human motion-capture data inputs (\cref{fig:teaser}).

\noindent This paper makes three contributions:\\
(i) We present \method{}, a kinematic retargeting pipeline that preserves the hand--object interaction rather than hand pose alone, pairing an interaction-graph Laplacian optimizer with a feed-forward network that amortizes the seed retargeting and is refined under object-aware contact losses.\\
(ii) We show that object awareness at the kinematic stage improves downstream RL, raising success from 20.6\% to 48.2\% over hand-only references. The learned retargeter transfers to an unseen dataset zero-shot or with light fine-tuning, without repeating optimization or amortization.\\
(iii) We surpass state-of-the-art (SOTA) on OakInk-v2~\cite{oakinkv2} and ARCTIC~\cite{arctic} with Allegro~\cite{allegrohand} and Inspire~\cite{inspirehand} hands, and achieve 46/56 successful hardware replays across 5 tasks.

%% file: sections/02_related_work.tex
\section{Related Work}
\label{sec:related}

\noindent\textbf{Learning manipulation from human data.}
Generalist robot policies~\cite{openvla,pi07} draw action supervision from large,
costly robot-trajectory datasets tied to particular hardware. 
Human data offers an abundant alternative and enters robot learning at two levels. At the representation level, it pretrains visual features, latent plans, or human--robot aligned embeddings~\cite{r3m,mimicplay,egobridge}, without ever mapping motion to the robot. At the action level, methods predict human actions that are retargeted to the robot~\cite{egovla}, or learn policies and controllers from reconstructed videos and kinematic human--object references~\cite{hop,videomanip,dextrack,bidexhd}, which first requires expressing the demonstration in the target embodiment. \method{} supplies that step, generating the object-aware references these pipelines consume.

\noindent\textbf{Retargeting from human data.}
\emph{Kinematic retargeting} maps human motion to robot configurations without physics.
Hand-centric DexPilot~\cite{dexpilot} and AnyTeleop~\cite{anyteleop} optimize keypoint
vectors, while GeoRT~\cite{geort} learns a direct map from human finger keypoints to
robot joints through geometric objectives. None constrains object-relative contact.
OmniRetarget~\cite{omniretarget} instead preserves body--object--terrain relations through an interaction mesh and sequential constrained optimization, but targets humanoid whole-body loco-manipulation.
\emph{Dynamic retargeting} enforces physical executability. ManipTrans~\cite{maniptrans}
pairs a hand-motion imitator with task-specific residual RL driven
by object-tracking and contact rewards. DexMachina~\cite{dexmachina} learns object-state tracking with RL under a curriculum
that progressively removes virtual object-control assistance, guided by explicit motion
and contact rewards. SPIDER~\cite{spider} instead uses sampling-based optimization,
refining simulated control sequences through annealed sampling with virtual contact
guidance. These recover physical plausibility, but receive an object-agnostic reference and must discover contacts themselves. \method{}
instead places the object in the kinematic stage so the dynamic stage starts from a reference carrying the demonstrated contact geometry.

Concurrent pipelines~\cite{toporetarget,c2dex,regrind} preserve hand--object structure with Laplacian objectives, but re-optimize each new frame or demonstration before RL. \method{} instead uses optimization only for training seed references. Its reusable retargeter is fine-tuned dataset-wide with object-aware losses to retarget new inputs without further optimization.

\noindent\textbf{Hand--object interaction.}
DexYCB~\cite{dexycb}, ARCTIC~\cite{arctic}, and OakInk-v2~\cite{oakinkv2} support
contact-aware reconstruction and synthesis. ContactOpt~\cite{contactopt} refines human pose
with predicted contact. Grasp'D~\cite{graspd} synthesizes human and robot grasps through
differentiable contact. ArtiGrasp~\cite{artigrasp} uses physics-based RL for bimanual
grasping and articulation. Unlike these reconstruction and synthesis methods, \method{}
uses contact attraction, contact direction, and penetration objectives for human-to-robot
trajectory retargeting.

%% file: sections/04_method.tex
\section{\method{}: Object-Aware Kinematic Retargeting}
\label{sec:method}

\begin{figure*}
    \centering
    \includegraphics[width=1.0\linewidth]{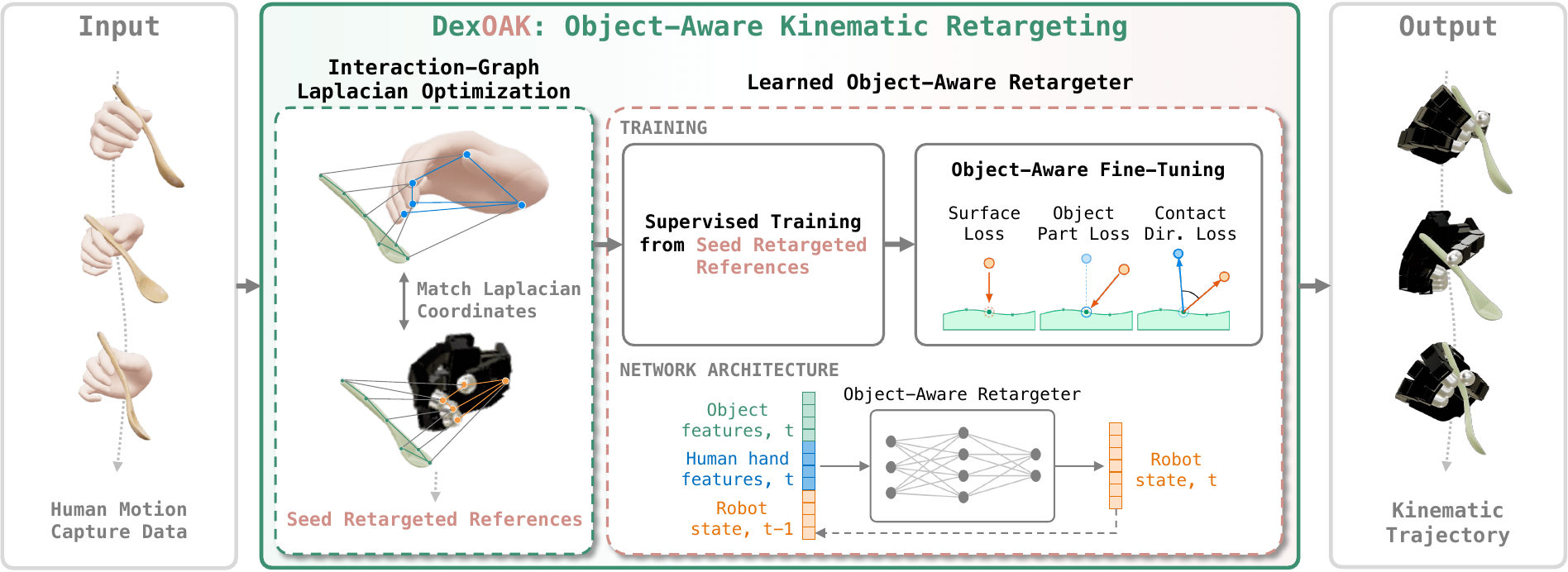}
    \caption{\textbf{Overview of \method{}.}
    From a human hand--object demonstration,
    our interaction-graph Laplacian optimizer matches human and robot Laplacian coordinates
    to produce a per-frame seed retargeted reference. A learned retargeter amortizes this
    optimization, autoregressively mapping the object cloud and human keypoints to the robot wrist and joints. It is trained on the seed references and further refined with
    object-aware losses. The resulting kinematic trajectory is later refined by an established residual RL stage (\cref{sec:rl}, omitted from the figure).
    Colors: object (green), human (blue), robot (orange).}
    \label{fig:method}
    \vspace{-0.4cm}
\end{figure*}

\method{} maps a $T$-frame human hand--object motion-capture sequence to a kinematic robot trajectory (\cref{fig:method}). At frame $t$, the inputs are $N_h$ indexed human hand keypoints $K_t=(k_t^i)_{i=1}^{N_h}\in\Real^{N_h\times3}$ from SMPL-X~\cite{smplx}, and an $N_p$-point object cloud $P_t\in\Real^{N_p\times3}$ sampled from the object meshes, both in the human wrist frame with pose $W_t\in\SEThree$. It outputs a robot wrist pose $w_t=(p_t,R_t)\in\SEThree$, with $p_t\in\Real^3$ and $R_t\in\SOthree$, and $J$ joint angles $q_t\in\mathcal Q=[q_{\mathrm{low}},q_{\mathrm{high}}]\subset\Real^{J}$. As $w_t$ is wrist-relative, the capture-frame robot wrist is $W_t\,w_t$, making the target invariant to global placement. We produce one trajectory per hand, coupling bimanual tasks only in the dynamic stage (\cref{sec:rl}).

Kinematic retargeting proceeds in two steps. An
interaction-graph Laplacian optimizer (left of \cref{fig:method}) builds paired
human--object and robot--object graphs, matches their Laplacian coordinates, and retargets
each frame into a seed retargeted reference (\cref{sec:interaction_graph}). 
A feed-forward network (right of \cref{fig:method}) amortizes these solutions (\cref{sec:supervised_amortization}) and
is then fine-tuned with object-aware contact losses
(\cref{sec:object_aware_finetuning}).
A separate stage (\cref{sec:rl}) then makes the trajectory physically executable.

\subsection{Interaction-Graph Laplacian Optimization}
\label{sec:interaction_graph}

\begin{figure*}[t]
  \centering
  \includegraphics[trim=0 0.7cm 0 0, clip, width=\textwidth]{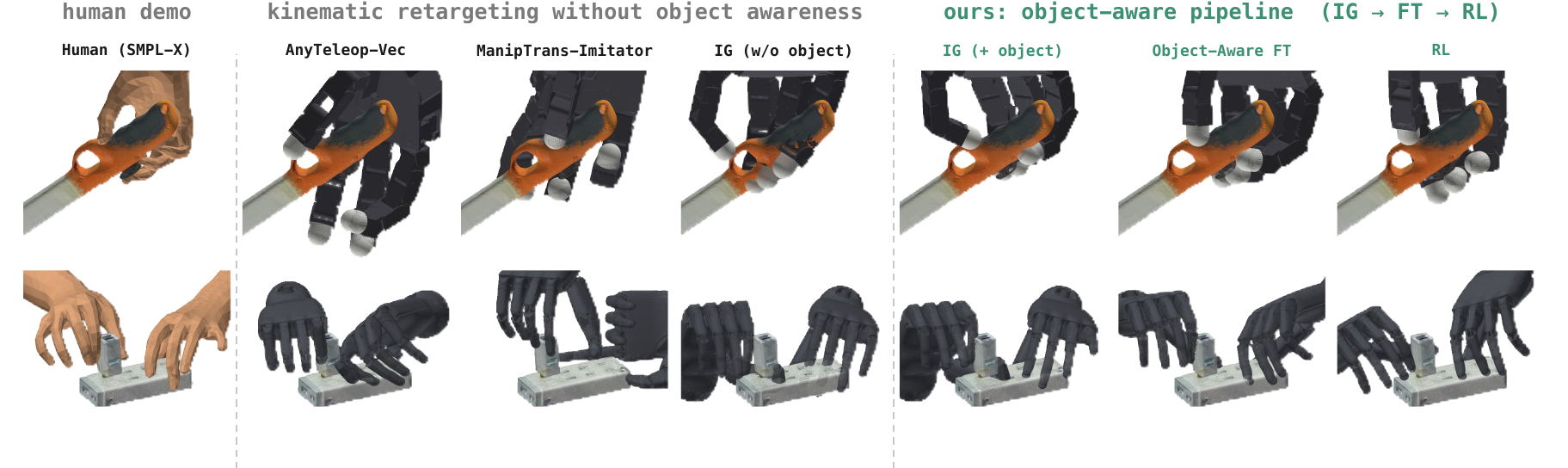}
  \caption{\textbf{\method{} stages and downstream RL recover fingertip contacts missed by prior methods.}
Two OakInk-v2~\cite{oakinkv2} task frames (top: ignite, Allegro hand~\cite{allegrohand};
bottom: unplug, Inspire hand~\cite{inspirehand}). Hand-only AnyTeleop-Vector~\cite{anyteleop}, ManipTrans imitator~\cite{maniptrans}, and
\emph{IG (w/o object)} optimizer leave fingertips off or inside the object.
\emph{IG (+ object)} optimizer, Object-Aware Fine-Tuning (FT), and RL (\cref{sec:rl}) progressively improve fingertip placement.
}
  \label{fig:contact-detail}
  \vspace{-0.4cm}
\end{figure*}

\noindent{\it Interaction Graph (IG).}
At frame $t$, we build two point sets. The source set
\(
    \mathcal S_t=\{k^1_t,\dots,k^{M_h}_t\}\cup\{o^1_t,\dots,o^{M_o}_t\}
\)
holds $M_h$ human hand keypoints (fingertips and wrist) from $K_t$, and
$M_o$ object anchors subsampled from $P_t$. 
The target set
\(
    \mathcal T_t(w_t,q_t)=\{r^1_t,\dots,r^{M_h}_t\}\cup\{o^1_t,\dots,o^{M_o}_t\}
\)
keeps the same object anchors and replaces the hand keypoints with robot
counterparts. Equal hand-node indices in $\mathcal S_t$ and $\mathcal T_t$ denote anatomical matches, such as thumb to thumb. Thus, $r^i_t=w_t\,\mathrm{FK}^i(q_t)$, where forward kinematics $\mathrm{FK}^i$ places robot
point $i$ in the robot root frame. Both sets have $N=M_h+M_o$ nodes. The shared object anchors make the Laplacian coordinates encode hand placement relative to the demonstrated object, unlike a hand-only graph (\emph{IG (w/o object)} vs. \emph{IG (+ object)} in \cref{fig:contact-detail}).

Following OmniRetarget~\cite{omniretarget}, we apply a 3D Delaunay tetrahedralization to $\mathcal S_t$. The neighborhood set $\mathcal N(i)$ contains the nodes connected to $i$ by a Delaunay edge and excludes $i$ itself. 
We reuse the same edges for the robot--object target $T_t$, making the robot reproduce the demonstrated neighborhood rather than form a new one. Anchor identities stay fixed across the sequence, while the tetrahedralization is recomputed each frame.
The Laplacian coordinate of a node $x_i$ is then defined as \(\Delta_i=\frac{1}{|\mathcal N(i)|}\sum_{j\in\mathcal N(i)}(x_i-x_j)\).
It measures the offset from the neighbor centroid and is invariant to global translation.

Our IG extends OmniRetarget's implementation in three ways. OmniRetarget represents
the hand by one wrist node and cannot resolve finger-level contact. 
Its mesh spans one object plus terrain, whereas ours pools all scene objects into one graph. It scales human keypoints to the robot by a height ratio, whereas we retain true scale relative to the object.

\noindent{\it Sequential per-frame optimization.}
We solve for robot wrist and joints that reproduce the human Laplacian coordinates under hard and soft constraints:
\begin{equation}
\begin{aligned}
    \min_{w_t,\,q_t}\;\;
    &\mathcal L_{\mathrm{IG}}+\mathcal L_{\mathrm{coll}}+\mathcal L_{\mathrm{temp}}\\
    \text{s.t.}\;\;
    &q_t\in\mathcal Q,\quad
     \lvert q_t-q^\star_{t-1}\rvert\preceq \mathbf v_{\max}\,\mathrm{d}t .
\end{aligned}
\label{eq:igraph}
\end{equation}
Superscript $\star$ denotes an optimizer solution.
The interaction-graph term is the mean squared difference of Laplacian coordinates,
$\mathcal L_{\mathrm{IG}}=\tfrac1N\sum_{i=1}^{N}
\lVert\Delta_i(\mathcal S_t)-\Delta_i(\mathcal T_t(w_t,q_t))\rVert_2^2$.
Collisions are computed as a hinge penalty,
$\mathcal L_{\mathrm{coll}}=\sum_{c}\max(0,\,-g_c(w_t,q_t))^2$,
where $g_c$ is the distance of collision pair $c$ minus its safety margin. Pairs cover robot--object and robot
self-collision, excluding adjacent links. Let $d_{\mathrm{rot}}$ denote the geodesic rotation distance on $\SOthree$. The temporal regularizer
$\mathcal L_{\mathrm{temp}}=\lVert q_t-q^\star_{t-1}\rVert_2^2
+\lVert p_t-p^\star_{t-1}\rVert_2^2
+d_{\mathrm{rot}}^2(R^\star_{t-1},R_t)$
keeps the joints and the wrist close to the previous frame solution.
Here $\mathbf v_{\max}\in\mathbb R^J$ gives per-joint velocity limits and $\preceq$ is elementwise. Wrist motion is only softly regularized by $\mathcal L_{\mathrm{temp}}$.
We solve the per-frame programs with sequential least-squares programming
(SLSQP)~\cite{pyslsqp}. Each program is
warm-started from the previous solution, giving the seed retargeted references
$\tau^\star=\{(w^\star_t,q^\star_t)\}_{t=0}^{T-1}$.

\subsection{Learned Object-Aware Retargeter}
\label{sec:learned_retargeter}

We amortize the optimizer into a feed-forward retargeter $f_\theta$ with autoregressive
conditioning, trained on the hand--object interactions and seed retargeted references
$\tau^\star$. The data spans many sequences and tasks, letting $f_\theta$ learn one shared retargeting function that generalizes to unseen data, instead of solving
each frame in isolation. It predicts the robot configuration from the
current observation and its own previous prediction,
$f_\theta\big(K_t,P_t,\hat q_{t-1},\hat w_{t-1}\big)=(\hat w_t,\hat q_t)$.

We represent each keypoint $k_t^i$ in $K_t$ by a 12D attribute, defined in the human wrist
frame: position (3), its SMPL-X~\cite{smplx} joint orientation as a 6D
rotation~\cite{zhou2019continuity} (6), and linear velocity (3). A PointNet-style
encoder~\cite{pointnet} applies a shared pointwise multilayer perceptron (MLP) and global max pooling to the object-point features. We concatenate the descriptor with the flattened keypoint attributes as the per-frame feature
$\phi_t$. The joint and wrist MLP decoders read $\phi_t$ concatenated with their respective predictions from the previous frame,
$\hat q_{t-1}$ or $\hat w_{t-1}$. This autoregressive conditioning mirrors the optimizer's warm start and ablations (\cref{tab:ablations}) support this design. We keep it free-running rather
than teacher-forced, so the network sees its own predictions during training and deployment, and condition the first frame on a zero-valued vector. The joint decoder passes its output through a per-joint sigmoid,
$\hat q_t=q_{\mathrm{low}}+\sigma(\cdot)\,(q_{\mathrm{high}}-q_{\mathrm{low}})$, keeping
$\hat q_t\in\mathcal Q$ without a joint-limit loss. The wrist decoder outputs
$\hat p_t\in\Real^3$ and a continuous, easier-to-regress 6D rotation~\cite{zhou2019continuity}, converted to $\hat R_t\in\SOthree$.

\subsubsection{Supervised Training}
\label{sec:supervised_amortization}

We first train $f_\theta$ autoregressively and fully supervised to reproduce the
optimization outputs $(w^\star_t,q^\star_t)$, with the per-frame loss
$\mathcal L_{\mathrm{sup}}=\mathcal L_q+\mathcal L_w$. It equally weights the $L_1$
joint-angle loss $\mathcal L_q$ and the wrist loss $\mathcal L_w$, an $L_1$ translation
term plus the geodesic $\SOthree$ rotation distance.
We write its output as
$\hat\tau^{\mathrm{sup}}=\{(\hat w^{\mathrm{sup}}_t,\hat q^{\mathrm{sup}}_t)\}_{t=0}^{T-1}$.
With the seed reference as its only target, this step reproduces $\tau^\star$ without improving it (\cref{tab:controlled_rl}). Its role is to initialize the object-aware fine-tuning in \cref{sec:object_aware_finetuning}, so those losses refine an existing solution rather than search for contacts from scratch. Without this initialization, fine-tuning
is too weakly constrained to converge (ablations in \cref{tab:ablations}).

\subsubsection{Object-Aware Fine-Tuning (FT)}
\label{sec:object_aware_finetuning}

The second step refines the same autoregressive network without labels. Matching Laplacian
coordinates preserves the hand--object neighborhood but does
not force the hand points onto the object surface, so contacts
can remain inaccurate
(\cref{fig:contact-detail}, \emph{IG (+ object)}). The objectives below act on these contacts
directly.

\noindent{\it Contact objectives.}
For any point $u$, let $o(u)$ be the nearest point of the object cloud $P_t$ and
$d(u)=\lVert u-o(u)\rVert_2$ its distance to the object surface. Following
\cref{sec:interaction_graph}, hand point $i$ has human position $k^i_t$ and robot
position $r^i_t$. We define three per-frame contact terms, illustrated in \cref{fig:method}:
\vspace{-2pt}
\begin{align}
    \mathcal L_{\mathrm{surf}} &=\textstyle\sum_{i}\gamma^i_t\,d(r^i_t)^2,
    \label{eq:loss_surf}\\
    \mathcal L_{\mathrm{part}} &=\textstyle\sum_{i}\gamma^i_t\lVert r^i_t-o(k^i_t)\rVert_2^2,
    \label{eq:loss_part}\\
    \mathcal L_{\mathrm{dir}} &=\textstyle\sum_{i}\gamma^i_t\big(1-\cos\theta^i_t\big),
    \label{eq:loss_dir}
\end{align}
\vspace{-2pt}
where $\theta^i_t$ is the angle between $r^i_t-o(k^i_t)$ and $k^i_t-o(k^i_t)$, and the
gate $\gamma^i_t$ activates a term only when the human point lies within a threshold
$\epsilon$ of the object surface. 
Gating matters because contact terms are informative only near the object and otherwise add noise during approach and release.
$\mathcal L_{\mathrm{surf}}$ pulls each robot point onto the object surface.
$\mathcal L_{\mathrm{part}}$ pulls it to $o(k^i_t)$, the nearest sampled object point to the human contact, so
each finger stays on the corresponding object part. $\mathcal L_{\mathrm{dir}}$ makes the robot
reach that point from the same side as the human, which separates opposite faces of thin
geometry.

\noindent{\it Regularization.}
Three further terms keep the prediction feasible and stable where the contact terms
give no signal. We reuse the collision penalty $\mathcal L_{\mathrm{coll}}$ from \cref{sec:interaction_graph}. In fact, $\mathcal L_{\mathrm{surf}}$ is symmetric about the surface, so non-penetration is handled by $\mathcal L_{\mathrm{coll}}$. $\mathcal L_{\mathrm{temp}}$ uses the same form as in the optimizer and compares the consecutive network predictions $(\hat w_{t-1},\hat q_{t-1})$ and $(\hat w_t,\hat q_t)$. The anchor term instead compares the prediction at frame $t$ with the supervised network output at the same frame
(\cref{sec:supervised_amortization}). It is
$\mathcal L_{\mathrm{anc}}=\lVert\hat q_t-\hat q^{\mathrm{sup}}_t\rVert_2^2
+\lVert\hat p_t-\hat p^{\mathrm{sup}}_t\rVert_2^2
+d_{\mathrm{rot}}^2(\hat R^{\mathrm{sup}}_t,\hat R_t)$.

The per-frame object-aware fine-tuning objective is the weighted sum:
\(\mathcal L_{\mathrm{ft}}=\sum_{\ell}\lambda_\ell\,\mathcal L_\ell\),
with $\ell\in\{\mathrm{surf},\mathrm{part},\mathrm{dir},\mathrm{coll},
\mathrm{anc},\mathrm{temp}\}$. The weights are tuned empirically and evaluated in \cref{sec:res_ablations}. The output is
the refined object-aware kinematic reference
$\hat\tau=\{(\hat w_t,\hat q_t)\}_{t=0}^{T-1}$. \cref{fig:contact-detail}, \emph{Object-Aware FT}, illustrates this stage's improvements.

\section{Physics-Based Dynamic Retargeting}
\label{sec:rl}
RL is well established for trajectory following in physics-based
character and humanoid control~\cite{deepmimic,beyondmimic}, and residual RL has since been used to correct base controllers for dexterous manipulation~\cite{maniptrans} and humanoid
loco-manipulation~\cite{resmimic}. We build on these methods for this stage.

For OakInk-v2 ablations (\cref{tab:controlled_rl}), we use the residual RL of ManipTrans~\cite{maniptrans}, and for ARCTIC generalization experiments (\cref{tab:generalization}), that of DexMachina~\cite{dexmachina}. Fixing RL within each table isolates the kinematic reference's contribution.
For the SOTA comparison across embodiments and datasets we use one unified
residual RL, combining DeepMimic tracking~\cite{deepmimic} with the ManipTrans
rewards~\cite{maniptrans} and the DexMachina object curriculum~\cite{dexmachina}. We
find that residual RL needs the object curriculum on ARCTIC, so we keep it on both
datasets and report a single protocol.

At step $t$, our unified residual policy observes simulator state $s_t$ and one-step-ahead targets from the kinematic reference $\hat\tau$ and demonstrated object state (pose and velocity) $x_t^o$.
It outputs a residual $\delta a_t=\pi(s_t,\hat\tau_{t+1},x_{t+1}^o)$ that corrects a base action $\bar a_t$ derived from $\hat\tau_{t+1}$, so the policy only has to fix what reference following cannot achieve.
The two are summed for the finger joints, $a_t=\bar a_t+\delta a_t$, and the actions are mapped onto the joint limits as
proportional-derivative (PD) position targets. 
The wrist is driven by a proportional-integral-derivative (PID) controller that maps position and orientation errors to a force and torque, respectively, to which the residual is added.
We train with reference-state initialization (RSI)~\cite{deepmimic} directly from $\hat\tau$ and the demonstrated object trajectory.
Unlike prior work~\cite{dexmachina,spider}, we do not pre-resolve its contacts in a
simulator before dynamic retargeting. We adapt the
domain-randomization ranges of the cited implementations (\cref{sec:exp_setup}), and
otherwise follow the ManipTrans and DexMachina
implementations.

%% file: sections/05_experiments.tex
\section{Experimental Results}
\label{sec:experiments}

\subsection{Experimental Setup}
\label{sec:exp_setup}

\paragraph{Datasets}
We use two hand--object motion-capture datasets, training on each and evaluating on
the subsets used by SOTA methods.
\textbf{OakInk-v2}~\cite{oakinkv2}: bimanual and single-hand object manipulation. We
train on the ManipTrans~\cite{maniptrans} set of 77 sequences across 42 objects (39
bimanual, 27 right-hand, 11 left-hand), reporting on the bimanual set for the SOTA
comparison (\cref{tab:sota}) and the right-hand set for the ablations
(\cref{tab:controlled_rl,tab:ablations}).
\textbf{ARCTIC}~\cite{arctic}: bimanual articulated-object manipulation.
We train on 30 sequences (5 objects, 6 clips each) and report on the 7-clip
subset used by DexMachina~\cite{dexmachina} and SPIDER~\cite{spider}
(\cref{tab:generalization,tab:sota}).

\paragraph{Methods compared}
We compare two groups of methods from \cref{sec:related}. \emph{Kinematic-reference} methods output a hand-centric, object-agnostic reference:
optimization-based AnyTeleop~\cite{anyteleop} in its
Position, Vector, and DexPilot~\cite{dexpilot} variants, and the learned GeoRT~\cite{geort}. 
Since each produces only a reference, we run each through the same dynamic-retargeting backend as ManipTrans~\cite{maniptrans} to isolate reference quality.
\emph{Full retargeting pipelines} are ManipTrans, DexMachina~\cite{dexmachina}, and
SPIDER~\cite{spider}. Only
ManipTrans contributes a kinematic retargeter, its imitator policy, making it our
primary comparison in \cref{tab:controlled_rl}. As it reports no kinematic
metrics, we score its imitator output ourselves.

\paragraph{Robot hands}
\textbf{Allegro}~\cite{allegrohand} is a four-finger hand
with 16 controllable DoF. \textbf{Inspire}~\cite{inspirehand} is a five-finger
hand with 6 controllable DoF and 6 mimic joints.

\paragraph{Kinematic metrics}
We score the kinematic reference with two metrics.
\textbf{Tip-position error} (TPE, mm) is the mean Euclidean distance between the robot and human fingertips.
\textbf{Contact completeness} (CC, \%) asks whether the robot consistently replicates human contact.
Human contact is defined by $d(k_t^i)<10$\,mm. Over frames with at least one such contact, CC is the fraction where every contacting robot fingertip lies within 10\,mm of the surface.
This deliberately conservative all-contact criterion requires the entire demonstrated contact pattern.
Both metrics are diagnostic rather than complete predictors of downstream success, but they track it closely across our object-aware stages (\cref{tab:controlled_rl}).

\paragraph{Dynamic retargeting metrics}
Prior methods use different success definitions, so we report each under its native metric.
\textbf{SR$_{\mathrm{MT}}$} is the strict
ManipTrans~\cite{maniptrans} criterion: a rollout succeeds when the
object-rotation, object-translation, joint-position, and fingertip errors fall below
$30^\circ$ (0.524\,rad), 3\,cm, 8\,cm, and 6\,cm. \textbf{SR$_{\mathrm{obj}}$} is the
object-only SPIDER~\cite{spider} criterion. It uses the same definitions as SR$_{\mathrm{MT}}$, but scores object motion alone, with different bounds of $28.6^\circ$ (0.5\,rad) and 10\,cm.
\textbf{ADD-AUC} is DexMachina's~\cite{dexmachina} object-tracking score, based on distances between object surface points under the achieved and demonstrated poses.

\paragraph{Implementation}
We solve \cref{eq:igraph} using $M_o{=}64$ object anchors, with 5\,mm robot--object and 8\,mm self-collision margins for $\mathcal L_{\mathrm{coll}}$. The retargeter uses $N_h{=}21$ keypoints per hand and an $N_p{=}1{,}024$-point object cloud. Its PointNet-style encoder applies shared pointwise ReLU MLP layers of widths 64, 128, and 256, followed by global max pooling. Its two MLP decoders each have three 512-unit hidden layers. We train with AdamW and a cosine schedule for 600 supervised epochs at learning rate (lr) $10^{-3}$ and 24 object-aware fine-tuning epochs at lr $10^{-5}$. The contact gate activates within $\epsilon{=}15$\,mm of the surface and was varied from 5 to 25\,mm with little effect on kinematic metrics. The fine-tuning weights $(\lambda_{\mathrm{surf}},\lambda_{\mathrm{part}},\lambda_{\mathrm{dir}},\lambda_{\mathrm{coll}},\lambda_{\mathrm{anc}},\lambda_{\mathrm{temp}})$ are $(5,10,1,10,1,1)$. Ablations use uniform $(1,1,1,1,1,1)$, contact-heavy $(50,100,20,10,1,1)$, and regularization-heavy $(5,10,1,100,50,10)$ variants (\cref{tab:ablations}). Residual RL uses 4{,}096 environments and PPO~\cite{schulman2017ppo}, following \cite{maniptrans,dexmachina}. We scale object friction by $[0.5,3]$ and mass by $\pm20\%$. Unless noted, results average 3 seeds in Tables~\ref{tab:controlled_rl}-\ref{tab:generalization} and 5 in \cref{tab:sota}, with each policy evaluated over 256 rollouts. Baseline results in Tables~\ref{tab:generalization} and \ref{tab:sota} are from the cited papers. All experiments use an NVIDIA H100.

\subsection{Ablations of Object Awareness}
\label{sec:res_controlled}

\begin{table}[t]
\centering
\small
\setlength{\tabcolsep}{3.6pt}
\renewcommand{\arraystretch}{1.05}
\caption{
\textbf{Kinematic-reference comparison on the Allegro hand (right-hand OakInk-v2).}
All rows use ManipTrans residual RL. Results are our reproductions. Best in \textbf{bold}.}
\label{tab:controlled_rl}
\resizebox{\linewidth}{!}{%
\begin{tabular}{@{}l ccc@{}}
\toprule
& \multicolumn{2}{c}{Kinematic metrics} & \multicolumn{1}{c}{RL metric} \\
\cmidrule(lr){2-3}\cmidrule(lr){4-4}
Kinematic reference & Contact compl.\,$\uparrow$ & TPE\,$\downarrow$ & SR$_{\mathrm{MT}}$\,$\uparrow$ \\
 & (\%) & (mm) & (\%) \\
\midrule
\multicolumn{4}{@{}l}{\textit{Without Object (Hand-Only Reference)}} \\
\quad AnyTeleop-Vector~\cite{anyteleop} & 3.5 & 59.5 & 18.3 \\
\quad AnyTeleop-Position~\cite{anyteleop} & 0.5 & 77.9 & 11.1 \\
\quad AnyTeleop-DexPilot~\cite{anyteleop, dexpilot} & 0.3 & 53.8 & 0.0 \\
\quad GeoRT~\cite{geort}                & 1.0 & 58.6 & 0.8 \\
\quad ManipTrans~\cite{maniptrans}      & 5.8 & 49.2 & 20.6 \\
\midrule
\multicolumn{4}{@{}l}{\textit{Object-Aware Reference (Ours)}} \\
\quad IG optimizer                       & 19.6 & 3.50 & 43.2 \\
\quad + supervised training         & 19.2 & 3.50 & 43.6 \\
\quad + Object-Aware FT (Full) & \textbf{21.9} & \textbf{3.21} & \textbf{48.2} \\
\bottomrule
\end{tabular}%
}
\end{table}

We compare object-aware and hand-only references on the OakInk-v2 Allegro right-hand tasks, all under the same ManipTrans residual RL (\cref{tab:controlled_rl}). Hand-only references reach at most 20.6\% RL success, with near-zero contact completeness and centimeter-scale tip errors. Object-aware references substantially improve both kinematic metrics and reach 48.2\% success, 2.3$\times$ the ManipTrans result. Placing fingers at demonstrated contacts lets residual RL refine dynamics rather than discover contacts. \cref{fig:contact-detail} illustrates the improvement from \emph{IG (w/o object)} to \emph{IG (+ object)}.

The object-aware rows in \cref{tab:controlled_rl} isolate our method's stages. The IG Laplacian optimizer supplies the reference, supervised training reproduces it, and object-aware fine-tuning improves every metric. \method{} adds 5.0 percentage points over the optimizer. The same ordering holds on ARCTIC (\cref{tab:generalization}: IG optimizer vs.\ \method{}). This supports our second contribution: the learned retargeter is differentiable, so we improve it under dataset-wide contact objectives rather than merely amortize the solver.

\subsection{Our Method Ablations}
\label{sec:res_ablations}

\begin{table}[t]
\centering
\small
\setlength{\tabcolsep}{4pt}
\renewcommand{\arraystretch}{1.05}
\caption{
\textbf{Component ablations of \method{} on the Allegro hand (right-hand OakInk-v2).}
Kinematic metrics only, without RL.
Each row changes one design choice from the full model, with all else fixed.}
\label{tab:ablations}
\begin{tabular}{@{}l cc@{}}
\toprule
\multirow{2}{*}{\textbf{Variant}}
 & Contact completeness\,$\uparrow$ & TPE\,$\downarrow$ \\
 & (\%) & (mm) \\
\midrule
\multicolumn{3}{@{}l}{\textit{Interaction-Graph Laplacian Optimizer}} \\
\quad without object anchors              & 2.8 & 41.5 \\
\quad without wrist node           & 19.6 & 3.51 \\
\midrule
\multicolumn{3}{@{}l}{\textit{Architecture and Training}} \\
\quad without object cloud               & 20.0 & 3.83 \\
\quad without autoregression & 16.1 & 9.29 \\
\quad single-sequence training       & 20.9 & 3.53 \\
\quad without supervised training  & 10.3 & 36.8 \\
\midrule
\multicolumn{3}{@{}l}{\textit{Object-Aware Losses (Leave-One-Out)}} \\
\quad without surface-contact            & 13.2 & 6.84 \\
\quad without object-part         & 19.2 & 7.71 \\
\quad without contact-direction          & 18.8 & 7.25 \\
\midrule
\multicolumn{3}{@{}l}{\textit{Loss-Weight Configurations}} \\
\quad uniform           & 19.2 & 5.03 \\
\quad contact-heavy     & 20.8 & 4.11 \\
\quad regularization-heavy   & 19.3 & 3.85 \\
\midrule
Full model                           & \textbf{21.9} & \textbf{3.21} \\
\bottomrule
\end{tabular}
\end{table}

\cref{tab:ablations} ablates our method's components on Allegro right-hand OakInk-v2 tasks (kinematic metrics only).
Removing any component degrades the kinematic metrics, justifying the design choices of \cref{sec:method}. Dropping the object anchors leaves a hand-only optimizer and yields 2.8\% contact completeness and 41.5\,mm TPE, showing that the object nodes carry the demonstrated contact geometry.
Skipping the supervised stage is almost as damaging (10.3\% and 36.8\,mm): without amortizing the optimizer first, the model has no usable initialization and fails to learn a useful retargeting from scratch.
Autoregression, the learned-network counterpart of the optimizer's warm start, is also important, helping smooth predictions.
Removing the object cloud from the network input costs little (20.0\% and 3.83\,mm), as object information still enters through the
losses.
All tested weightings remain close to the full model and substantially outperform the hand-only references from \cref{tab:controlled_rl}. The benefit thus comes from the
loss terms themselves rather than their precise balance.

\subsection{Generalization}
\label{sec:res_generalization}

We test whether our retargeter transfers across datasets without re-optimization. We evaluate on ARCTIC with the Allegro hand under DexMachina's RL (\cref{tab:generalization}). We report speed and ADD-AUC. Speed is calculated as evaluated frames divided by total target-data kinematic-stage wall-clock time, including inference, per-frame optimization, and any one-time training or fine-tuning, excluding RL.
The top block trains and evaluates \method{} on ARCTIC, following our typical pipeline and providing baselines for the generalization rows below. In the bottom block, our OakInk-v2 \method{} model is applied zero-shot to the ARCTIC data, or after object-aware fine-tuning on ARCTIC (\emph{+ Object-Aware FT}). Neither transfer setting requires per-frame optimization.

Zero-shot, the OakInk-v2 model reaches 87.9 ADD-AUC, above DexMachina at 83.4, and
generates references 7 times faster than our in-domain optimizer (108.9 vs.\ 15.5\,Hz).
Object-aware fine-tuning on ARCTIC raises ADD-AUC to 90.0, near in-domain
\method{} at 91.3. 
Thus, our trained retargeter generalizes zero-shot to new datasets, scaling \method{} to data it was never optimized on. Moreover, target-data object-aware fine-tuning improves ADD-AUC while remaining far faster than the in-domain \method{} (\emph{+ Object-Aware FT} vs. \emph{Ours: \method{}} in \cref{tab:generalization}).

\begin{table}[t]
\centering
\small
\setlength{\tabcolsep}{4pt}
\renewcommand{\arraystretch}{1.05}
\caption{
\textbf{Cross-dataset transfer of \method{} on the Allegro hand (bimanual OakInk-v2 $\rightarrow$ ARCTIC).}
All rows use DexMachina RL. Speed is evaluated frames divided by total target-data kinematic-stage time, including any optimization, training, or fine-tuning but excluding RL. ADD-AUC measures dynamic success. Best in \textbf{bold}, second-best \underline{underlined}.}
\label{tab:generalization}
\resizebox{\linewidth}{!}{%
\begin{tabular}{@{}l c @{\hspace{5pt}\vrule width 0.3pt\hspace{5pt}} c@{}}
\toprule
Method & Speed (Hz)\,$\uparrow$ & ADD-AUC (\%)\,$\uparrow$ \\
\midrule
\multicolumn{3}{@{}l}{\textit{In-domain on ARCTIC}} \\
\quad DexMachina~\cite{dexmachina}       & 13.4              & 83.4  \\
\quad Ours: IG optimizer                 & 15.5              & 88.9  \\
\quad Ours: \method{}               & 1.4               & \textbf{91.3}  \\
\midrule
\multicolumn{3}{@{}l}{\textit{Ours: generalization OakInk-v2 $\rightarrow$ ARCTIC}} \\
\quad Zero-shot                          & \textbf{108.9}     & 87.9  \\
\quad + Object-Aware FT on ARCTIC         & \underline{57.2}   & \underline{90.0}  \\
\bottomrule
\end{tabular}%
}
\end{table}

\begin{table}[!t]
\centering
\caption{%
\textbf{\method{}+RL vs.\ SOTA retargeting.} Methods use their own dynamic refinement, while ours uses the residual RL of \cref{sec:rl}. --: not reported. $^{*}$ SPIDER uses more OakInk-v2 sequences, making its comparison indicative. Best in \textbf{bold}.}
\label{tab:sota}
\setlength{\tabcolsep}{3.8pt}
\renewcommand{\arraystretch}{1.05}
\small
\resizebox{\linewidth}{!}{%
\begin{tabular}{@{}l ccc@{}}
\toprule
Method & SR$_{\mathrm{MT}}$ (\%)\,$\uparrow$ & SR$_{\mathrm{obj}}$ (\%)\,$\uparrow$ & ADD-AUC (\%)\,$\uparrow$ \\
\midrule
\multicolumn{4}{@{}l}{\textit{OakInk-v2 (bimanual), Inspire hand}} \\
\quad ManipTrans~\cite{maniptrans} & 39.5 & -- & -- \\
\quad SPIDER~\cite{spider}         & --    & 47.9$^{*}$ & -- \\
\quad \textbf{\method{}+RL}       & \textbf{48.4} & \textbf{58.7} & \textbf{55.1} \\
\midrule
\multicolumn{4}{@{}l}{\textit{OakInk-v2 (bimanual), Allegro hand}} \\
\quad SPIDER~\cite{spider}         & -- & 45.9$^{*}$ & -- \\
\quad \textbf{\method{}+RL}       & \textbf{58.6} & \textbf{64.1} & \textbf{61.5} \\
\midrule
\multicolumn{4}{@{}l}{\textit{ARCTIC, Inspire hand}} \\
\quad DexMachina~\cite{dexmachina} & -- & 67.1 & 62.7 \\
\quad SPIDER~\cite{spider}         & -- & 42.0 & -- \\
\quad \textbf{\method{}+RL}       & \textbf{58.3} & \textbf{83.4} & \textbf{77.9} \\
\midrule
\multicolumn{4}{@{}l}{\textit{ARCTIC, Allegro hand}} \\
\quad DexMachina~\cite{dexmachina} & -- & -- & 83.4 \\
\quad \textbf{\method{}+RL}       & \textbf{67.2} & \textbf{97.3} & \textbf{92.1} \\
\bottomrule
\end{tabular}%
}
\end{table}

\subsection{Comparison to State of the Art}
\label{sec:res_sota}

\begin{figure*}[t]
    \centering
    \includegraphics[width=\textwidth]{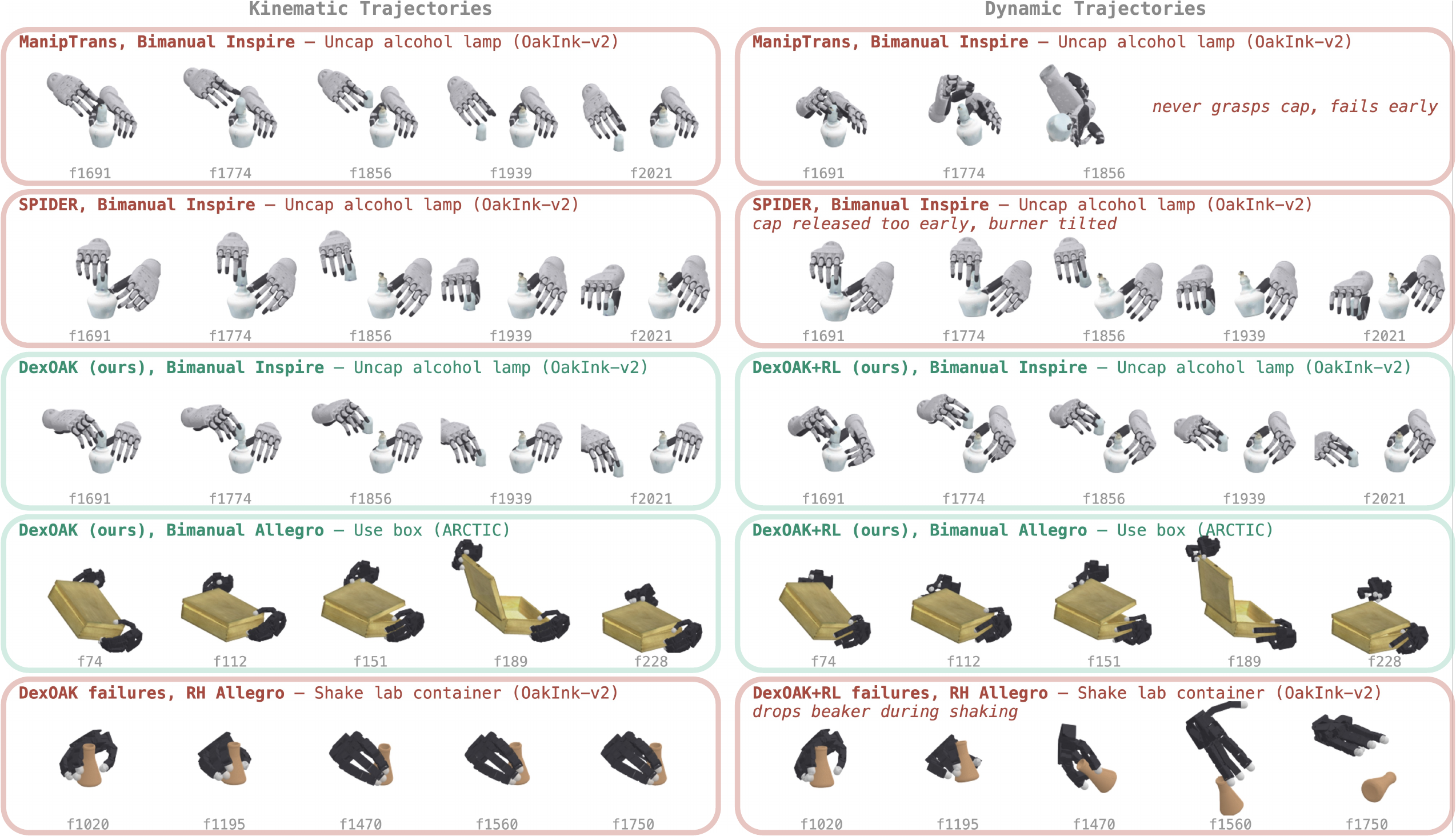}
    \caption{\textbf{Qualitative results across embodiments and datasets.}
Kinematic references (left) and dynamically refined trajectories (right). The top three rows compare ManipTrans~\cite{maniptrans}, SPIDER~\cite{spider}, and \method{}(+RL) on \emph{uncap alcohol lamp} (OakInk-v2~\cite{oakinkv2}, Inspire~\cite{inspirehand}). The remaining rows show \method{}(+RL) on \emph{use box} (ARCTIC~\cite{arctic}, Allegro~\cite{allegrohand}) and a failure on \emph{shake lab container} (OakInk-v2, Allegro), where the grasp breaks during shaking.}
    \label{fig:qualitative}
    \vspace{-0.4cm}
\end{figure*}

\cref{tab:sota} compares our \method{}+RL pipeline to ManipTrans, DexMachina, and
SPIDER on both hands over bimanual OakInk-v2 and ARCTIC, each under its own dynamic
retargeting and native metric. \emph{Ours} uses the
residual RL of \cref{sec:rl}, so the ARCTIC numbers here are not directly comparable
to \cref{tab:generalization}, which runs DexMachina's RL. Each baseline runs its published
pipeline, so this table compares complete systems rather than isolating the kinematic
reference, which \cref{tab:controlled_rl,tab:generalization} establish under a fixed
backend. 
On ARCTIC with the Inspire
hand, \method{}+RL doubles SPIDER's object success, 83.4 against 42.0
SR$_{\mathrm{obj}}$, and adds more than 15 ADD-AUC points to DexMachina to reach
77.9. The ordering is the same on OakInk-v2, where \method{}+RL outperforms ManipTrans
under SR$_{\mathrm{MT}}$ and SPIDER under SR$_{\mathrm{obj}}$. Across both hands, \method{}+RL outperforms each reported SOTA method under its native metric.
Interestingly, RL success
is higher with Allegro than Inspire (67.2 vs.\ 58.3 SR$_{\mathrm{MT}}$ on ARCTIC), despite its bulkier fingers.
This may reflect Allegro’s 16 controllable DoF versus Inspire’s 6, allowing Allegro to cover a wider
range of human hand configurations.

Qualitative results (\cref{fig:qualitative}) mirror \cref{tab:sota}. On the OakInk-v2 task
\emph{uncap alcohol lamp}, dynamic retargeting fails for ManipTrans and SPIDER, while
\method{}+RL executes the full motion. Nevertheless, failures remain. 
On highly dynamic tasks (e.g., \emph{shake lab container}), the grasp breaks mid-trajectory because kinematic retargeting lacks knowledge of sequence dynamics and the residual policy alone must supply momentum control.
Further examples are in the supplementary video.

\subsection{Real-Robot Experiments}
\label{sec:res_real}

These experiments evaluate the \method{}+RL pipeline for generating executable robot trajectories on 5 OakInk-v2~\cite{oakinkv2} tasks. For each task--method pair, we evaluate 3 trained policies (one per seed).
From each policy, we retain the first rollout satisfying $SR_{\mathrm{MT}}$ in a fixed evaluation order, record its wrist poses and finger joint angles, and replay it open-loop 4 times on a UR5 + right Allegro setup, yielding up to 12 trials per task (fewer when some policies produce no successful rollout).
We replay the 60\,Hz rollout in joint space, obtaining UR5 targets by IK from the recorded wrist poses and sending joint-position targets to both the UR5 and Allegro.
Before each replay, the object is manually reset to its marked initial pose, within $5\,\mathrm{mm}$ of the simulated initialization. Object dimensions match their mocap counterparts within \(2\,\mathrm{mm}\).
A replay succeeds if it completes the demonstrated object motion:
object transport to the demonstrated final position, gate/laptop closure,
bottle tilt-and-return, plug removal, knife cutting-stroke-and-return,
or button press. In \cref{tab:real_robot}, brackets report seed-averaged simulation $SR_{\mathrm{MT}}$ (how often the policy succeeds in simulation), while $x/y$ reports hardware repeatability of the selected rollouts.
Hardware success is therefore conditional on simulation success. We compare with ManipTrans~\cite{maniptrans}, trained with its original recipe.

\method{}+RL achieves near-perfect real-robot performance on the first three tasks: \emph{Take out}, \emph{Close gate} and \emph{Pour}. 
On \emph{Unplug} and \emph{Cut}, abrupt or high-frequency motions can cause the plug or knife to slip from the grasp, exacerbated by open-loop sensitivity to initialization and consistent with the dynamic failures observed in simulation (\cref{fig:qualitative}). Low simulation success further limits replayable \emph{Cut} trajectories.
Compared with ManipTrans, \method{}+RL
improves $SR_{\mathrm{MT}}$ on all 5 tasks, and matches or substantially exceeds its
available hardware success, with the clearest gain on \emph{Unplug}.

We additionally apply the frozen \method{} retargeter zero-shot to two tasks held out from all retargeter training: \emph{Close Laptop} and \emph{Press Button}, followed by the same task-specific RL stage. Both are non-prehensile interactions with objects from unseen categories. They achieve \(99\%\) and \(100\%\) simulation success, respectively, and \(10/12\) hardware successes each.
These results show that \method{} generalizes to unseen interactions, producing references that RL can refine into replayable hardware trajectories. Representative examples are shown in \cref{fig:teaser} and the supplementary video.

\newcommand{\hwsr}[2]{#1\,{\scriptsize[#2\%]}}

\begin{table}[t]
\caption{
\textbf{Real-robot results.}
The first successful simulation rollout per policy (3 seeds) is replayed
4 times on a UR5 + right Allegro. Entries show hardware successes/trials [simulation \(SR_{\mathrm{MT}}\)]. Denominators \(<\)12: fewer than 3 policies yielded a successful rollout. --: no successful rollout.
}
\label{tab:real_robot}
\centering
\resizebox{\linewidth}{!}{%
\begin{tabular}{@{}lccccc@{}}
\toprule
Method
& Take out
& Close gate
& Pour
& Unplug
& Cut \\
\midrule

\mbox{\method{}+RL}
& \textbf{\hwsr{12/12}{90}}
& \textbf{\hwsr{12/12}{41}}
& \textbf{\hwsr{10/12}{95}}
& \textbf{\hwsr{8/12}{93}}
& \textbf{\hwsr{4/8}{3}} \\

ManipTrans~\cite{maniptrans}
& \hwsr{6/12}{83}
& \hwsr{\textbf{12/12}}{12}
& \hwsr{8/12}{16}
& \hwsr{0/4}{1}
& \hwsr{--}{0} \\

\bottomrule
\end{tabular}%
}
\end{table}

%% file: sections/08_conclusion.tex
\section{Conclusion}
\label{sec:conclusion}
We introduce \method{}, an object-aware kinematic framework that retargets human
hand--object demonstrations into dexterous robot trajectories. It preserves object-relative
interaction geometry through interaction-graph Laplacian optimization, amortizes it
with a feed-forward retargeter refined by object-aware contact losses, and uses its output as the base action for residual RL. Across single-hand and bimanual tasks on two robot
embodiments, \method{}+RL clearly outperforms SOTA retargeting methods. Its RL-refined trajectories also replay successfully on real hardware.

Future work could share dynamics corrections across tasks and hands through one task- and embodiment-conditioned residual policy, and reach beyond motion-capture by retargeting hand--object trajectories reconstructed from RGB video. Both steps would extend \method{} into an engine that turns human observation into executable dexterous robot data.

%% file: sections/supplementary.tex
\setcounter{section}{0}
\renewcommand{\thesection}{S\arabic{section}}
\setcounter{table}{0}
\renewcommand{\thetable}{S\arabic{table}}
\setcounter{figure}{0}
\renewcommand{\thefigure}{S\arabic{figure}}

\section*{Supplementary Material}
This supplement gives the implementation details, hyperparameters, exact
sequence lists, and additional results omitted from the main paper for space. It
follows the same order as the paper, from preprocessing and the two kinematic stages
to the dynamic-retargeting stage.

\section{Data preprocessing}
\label{sec:supp_preproc}

\paragraph{Kinematic retargeting}
We follow the ManipTrans~\cite{maniptrans} data pipeline. From each raw
motion-capture sequence we load the SMPL-X~\cite{smplx} hand parameters, and extract $21$ hand keypoints per hand: $16$ articulation joints plus $5$ fingertip mesh vertices. OakInk-v2 is
subsampled from $120$ to $60$\,Hz. For each manipulated object we load its scan mesh
and take a fixed $1024$-point cloud by a deterministic vertex stride. Hands and object
are expressed in the per-frame human wrist frame. Sequences are cut into manipulation primitives following the DexManipNet
segmentation (\cref{tab:supp_oakink}): bimanual clips use the frame interval where both
hands are active, single-hand clips that hand's interval. We then assemble per-frame
features, a $12$-D vector per keypoint (position, 6D rotation~\cite{zhou2019continuity},
and velocity in the wrist frame) and the object cloud. Contact geometry is enforced downstream by the interaction-graph optimizer and the contact
losses (\cref{sec:method}).

\paragraph{Dynamic retargeting}
We follow ManipTrans~\cite{maniptrans} for the physics stage. Each object is staged
into a single free-body URDF with a convex-decomposition collision mesh and inertial
parameters from its hull. The kinematic reference $\hat\tau$ is exported per hand into
the reference format the simulator consumes, a floating $6$-DoF wrist chain plus the
finger joints (mimic joints excluded), the fingertip keypoints, and the object pose,
all in metres. No temporal resampling is applied. The reference is used in two ways during
training: as position targets for reference-state initialization and control, and as
the target for the imitation rewards.

\section{Interaction-graph node sets}
\label{sec:supp_graph}

Here we specify the exact points used to build the interaction graph
(\cref{sec:interaction_graph}) for each embodiment. The source set $\mathcal S_t$
uses $M_h$ human hand keypoints and $M_o$ object anchors, and the target set
$\mathcal T_t$ replaces the human keypoints with anatomically matched robot points
while keeping the same anchors.

\paragraph{Human keypoints}
The human hand is represented as $21$ keypoints in SMPL-X ordering (index $0$ is the
wrist, indices $\{4,8,12,16,20\}$ are the thumb, index, middle, ring, and pinky
fingertips). The graph uses only the wrist and the fingertips matched to the robot, so
$M_h$ equals the number of robot fingertips plus the wrist.

\paragraph{Robot nodes}
Each robot node is the tip frame of the corresponding finger, matched thumb-to-thumb,
index-to-index, and so on, plus a wrist node at the hand base. \cref{tab:supp_graph}
lists the exact link frames per hand. The Allegro hand has four fingers, so the human
pinky (keypoint $20$) has no counterpart and is dropped from the source set. The
Inspire hand has five fingers and matches all five human fingertips. The object anchors
are common to both sets, $M_o{=}64$ points subsampled from the object cloud by a
deterministic uniform stride, and the graph edges are a 3D Delaunay tetrahedralization
built once on the source nodes and reused on the target (\cref{sec:interaction_graph}).

\begin{table}[t]
\centering
\small
\setlength{\tabcolsep}{4pt}
\renewcommand{\arraystretch}{1.15}
\caption{\textbf{Interaction-graph nodes per embodiment.} Robot tip frames matched to
MANO fingertip keypoints, plus a shared wrist node. Allegro drops the human pinky.}
\label{tab:supp_graph}
\begin{tabular}{@{}llcc@{}}
\toprule
Finger & MANO kpt & Allegro tip frame & Inspire tip frame \\
\midrule
Thumb  & 4  & \texttt{link\_15.0\_tip} & \texttt{thumb\_tip}  \\
Index  & 8  & \texttt{link\_3.0\_tip}  & \texttt{index\_tip}  \\
Middle & 12 & \texttt{link\_7.0\_tip}  & \texttt{middle\_tip} \\
Ring   & 16 & \texttt{link\_11.0\_tip} & \texttt{ring\_tip}   \\
Pinky  & 20 & \textemdash             & \texttt{pinky\_tip}  \\
Wrist  & 0  & \texttt{wrist}           & \texttt{hand\_base\_link} \\
\midrule
\multicolumn{2}{@{}l}{Nodes $M_h$ (incl.\ wrist)} & 5 & 6 \\
\bottomrule
\end{tabular}
\end{table}

\section{Dynamic-retargeting implementation}
\label{sec:supp_rl}

We expand the residual-RL stage of \cref{sec:rl}. The policy is trained with PPO in a
GPU-parallel simulator on the Allegro hand ($16$ finger DoF) and the Inspire hand ($6$ finger DoF) with a floating $6$-DoF
wrist.

\paragraph{Observation}
At step $t$ the policy observes three groups. Proprioception gives the finger joint
angles with their sine and cosine and the wrist state. The object state gives its pose
relative to the wrist, its linear and angular velocity, and the per-fingertip contact
forces. The one-step-ahead targets, read from the kinematic reference
$\hat\tau_{t+1}$, give the wrist pose, finger joints, and object pose and velocity,
together with their residual to the current state.

\paragraph{Action and control}
The policy outputs a residual $\delta a_t$ that corrects a base action $\bar a_t$
derived from $\hat\tau_{t+1}$. For the finger joints the two are summed,
$a_t=\bar a_t+\delta a_t$, mapped onto the joint limits, and tracked by
proportional-derivative position control. For the wrist,
the residual is used to apply a force and torque command to the floating-wrist joints via a PID controller.

\paragraph{Reference-state initialization}
Unlike prior work that first refines the references in simulation, we initialize
directly from $\hat\tau$. Each episode starts from a frame drawn uniformly in the first
$98\%$ of the sequence, with the simulator joint, wrist, and object states set to the
reference at that frame.

\paragraph{Rewards and curriculum}
The rewards follow ManipTrans~\cite{maniptrans}, tracking the reference wrist pose,
finger joints, and object pose and velocity, with success tolerances tightened
exponentially over training. For the generalization and SOTA runs we add
the DexMachina~\cite{dexmachina} object curriculum, which relaxes object gravity,
friction, and control assistance early in training and anneals them to the true
dynamics. All observations, rewards, and PPO settings otherwise follow the cited
backends (\cref{tab:supp_hparams_rl}).

\section{Hyperparameters}
\label{sec:supp_hparams}

\cref{tab:supp_hparams_kin} lists the kinematic-stage settings and
\cref{tab:supp_hparams_rl} the RL stage. The six object-aware fine-tuning loss weights
are those of \cref{tab:ablations}.

\begin{table}[t]
\centering
\small
\setlength{\tabcolsep}{5pt}
\renewcommand{\arraystretch}{1.1}
\caption{\textbf{Kinematic-stage hyperparameters.} Optimizer, network, and the two
training stages. Decoder width and depth differ by embodiment.}
\label{tab:supp_hparams_kin}
\begin{tabular}{@{}ll@{}}
\toprule
\multicolumn{2}{@{}l}{\textit{Interaction-graph optimizer (SLSQP)}} \\
Object anchors $M_o$            & 64 \\
Object cloud size              & 1024 \\
Adjacency                      & Delaunay tetra (kNN fallback, $k{=}20$) \\
Joint velocity limit $v_{\max}$ & 12\,rad/s \\
Frame step $\mathrm{d}t$        & $1/30$\,s \\
Max iterations / tol           & 64 / $10^{-6}$ \\
Robot-object margin            & 2\,mm \\
Self-collision margin          & 8\,mm \\
\midrule
\multicolumn{2}{@{}l}{\textit{Network}} \\
Object encoder                 & PointNet $3{\to}64{\to}128{\to}256$, max-pool \\
Keypoint tokens                & $2{\times}21{=}42$, 12-D each \\
Decoder (Allegro)              & 8 layers, width 1024 \\
Decoder (Inspire)              & 4 layers, width 512 \\
Wrist head                     & translation (3) + 6D rotation (6) \\
Window length                  & 9 frames, autoregressive \\
\midrule
\multicolumn{2}{@{}l}{\textit{Supervised training}} \\
Optimizer / schedule           & AdamW / cosine (min-lr $10^{-7}$) \\
Learning rate                  & $10^{-3}$ \\
Epochs / batch                 & 600 / 128 \\
\midrule
\multicolumn{2}{@{}l}{\textit{Object-aware fine-tuning}} \\
Learning rate                  & $10^{-5}$ \\
Epochs / batch                 & 25 / 128 \\
Contact gate $\epsilon$        & 15\,mm \\
Non-penetration                & radius-aware, summed per frame \\
Loss weights $\lambda_\ell$    & see \cref{tab:ablations} \\
\bottomrule
\end{tabular}
\end{table}

\begin{table}[t]
\centering
\small
\setlength{\tabcolsep}{5pt}
\renewcommand{\arraystretch}{1.1}
\caption{\textbf{Dynamic-retargeting (RL) hyperparameters.} Residual PPO on the Allegro
hand.}
\label{tab:supp_hparams_rl}
\begin{tabular}{@{}ll@{}}
\toprule
Parallel environments          & 4096 \\
Simulation step                & $1/60$\,s \\
Episode length                 & 1200 steps \\
Finger control                 & PD position \\
Action dimension               & $6+16$ (wrist + fingers) \\
Discount $\gamma$              & 0.99 \\
GAE $\lambda$                  & 0.95 \\
Clip $\varepsilon$             & 0.2 \\
Rollout horizon                & 32 \\
Minibatch size                 & 1024 \\
Epochs per update              & 5 \\
Gradient-norm clip             & 1.0 \\
Learning rate                  & 5e-4 \\
Hardware                       & 1 $\times$ NVIDIA H100 \\
\bottomrule
\end{tabular}
\end{table}

\section{Datasets and sequence lists}
\label{sec:supp_seqs}

\begin{table}[t]
\centering
\small
\setlength{\tabcolsep}{5pt}
\caption{\textbf{ARCTIC subset used} (7 clips, 5 objects). Clip string is
\texttt{object,start,end,subject,use}, over seeds $\{42,24,66,15,113\}$.}
\label{tab:supp_arctic}
\resizebox{\linewidth}{!}{%
\begin{tabular}{@{}llc@{}}
\toprule
Clip & Sequence & Frames \\
\midrule
\texttt{box30-230-s01-u01}        & \texttt{box\_use\_01}       & 30--230 \\
\texttt{ketchup30-130-s01-u01}    & \texttt{ketchup\_use\_01}   & 30--130 \\
\texttt{ketchup40-340-s01-u02}    & \texttt{ketchup\_use\_02}   & 40--340 \\
\texttt{mixer30-200-s01-u01}      & \texttt{mixer\_use\_01}     & 30--200 \\
\texttt{mixer40-340-s01-u01}      & \texttt{mixer\_use\_01}     & 40--340 \\
\texttt{notebook40-340-s02-u02}   & \texttt{notebook\_use\_02}  & 40--340 \\
\texttt{waffleiron40-340-s01-u01} & \texttt{waffleiron\_use\_01} & 40--340 \\
\bottomrule
\end{tabular}%
}
\end{table}

\begin{table}[h!]
\centering
\small
\setlength{\tabcolsep}{5pt}
\caption{\textbf{ARCTIC training sequences} (30 clips, 5 objects), used only for
training. Evaluation uses the 7-clip subset of \cref{tab:supp_arctic}. Sequence name is
\texttt{subject\_\_object\_action\_index}.}
\label{tab:supp_arctic_train}
\resizebox{\linewidth}{!}{%
\begin{tabular}{@{}lll@{}}
\toprule
\texttt{s01\_\_box\_use\_01}   & \texttt{s07\_\_mixer\_use\_01}    & \texttt{s04\_\_notebook\_use\_01} \\
\texttt{s02\_\_box\_use\_01}   & \texttt{s01\_\_mixer\_grab\_01}   & \texttt{s06\_\_notebook\_use\_01} \\
\texttt{s04\_\_box\_use\_01}   & \texttt{s01\_\_ketchup\_use\_01}  & \texttt{s07\_\_notebook\_use\_01} \\
\texttt{s06\_\_box\_use\_01}   & \texttt{s01\_\_ketchup\_use\_02}  & \texttt{s04\_\_notebook\_grab\_01} \\
\texttt{s07\_\_box\_use\_01}   & \texttt{s02\_\_ketchup\_use\_01}  & \texttt{s01\_\_waffleiron\_use\_01} \\
\texttt{s01\_\_box\_grab\_01}  & \texttt{s04\_\_ketchup\_use\_02}  & \texttt{s02\_\_waffleiron\_use\_01} \\
\texttt{s01\_\_mixer\_use\_01} & \texttt{s06\_\_ketchup\_use\_01}  & \texttt{s04\_\_waffleiron\_use\_01} \\
\texttt{s02\_\_mixer\_use\_01} & \texttt{s01\_\_ketchup\_grab\_01} & \texttt{s06\_\_waffleiron\_use\_01} \\
\texttt{s04\_\_mixer\_use\_01} & \texttt{s01\_\_notebook\_use\_01} & \texttt{s07\_\_waffleiron\_use\_01} \\
\texttt{s06\_\_mixer\_use\_01} & \texttt{s02\_\_notebook\_use\_02} & \texttt{s01\_\_waffleiron\_grab\_01} \\
\bottomrule
\end{tabular}%
}
\end{table}

We train on the full sets described in \cref{sec:exp_setup} and report on the
prior-work subsets. \cref{tab:supp_oakink} lists the 39 bimanual tasks (Inspire), the 27 right-hand tasks
(Allegro), and the 11 left-hand tasks (Allegro, training only). Each is identified by its sequence hash and segment index
with its action label and manipulated object. For ARCTIC, \cref{tab:supp_arctic} lists
the 7-clip evaluation subset and \cref{tab:supp_arctic_train} the 30 training
sequences.

\begin{table*}[t]
\centering
\small
\setlength{\tabcolsep}{5pt}
\renewcommand{\arraystretch}{1.1}
\caption{\textbf{OakInk-v2 tasks used.} Identifier is \texttt{seqhash@segment}.
Bimanual tasks are evaluated on Inspire, right-hand on Allegro, and the left-hand
tasks are used for training only. Object and action labels are abbreviated.}
\label{tab:supp_oakink}
\begin{tabular}{@{}lll@{\hspace{1.1em}}lll@{\hspace{1.1em}}lll@{}}
\toprule
\multicolumn{3}{c}{\textit{Bimanual (Inspire), 39}} &
\multicolumn{3}{c}{\textit{Right-hand (Allegro), 27}} &
\multicolumn{3}{c}{\textit{Left-hand (Allegro), 11, train}} \\
\cmidrule(lr){1-3}\cmidrule(lr){4-6}\cmidrule(lr){7-9}
ID & Action & Object & ID & Action & Object & ID & Action & Object \\
\midrule
\texttt{c9cd3@0} & scoop & spoon & \texttt{c9cd3@1} & spread & spoon & \texttt{e76b2@1} & hold & apple \\
\texttt{d78fa@0} & open gate & microwave & \texttt{d7c33@1} & rearrange & spoon & \texttt{155cb@4} & place rack & test tube \\
\texttt{e76b2@0} & cut & knife & \texttt{e76b2@2} & place in & apple & \texttt{81a95@3} & rearrange & tripod \\
\texttt{e76b2@12} & open gate & microwave & \texttt{e76b2@4} & pour & bottle & \texttt{81a95@4} & place mesh & mesh \\
\texttt{e76b2@13} & take out & bowl & \texttt{5ced9@5} & remove lid & box lid & \texttt{81a95@5} & heat beaker & beaker \\
\texttt{e76b2@14} & close gate & microwave & \texttt{fb969@4} & rearrange & mouse & \texttt{c50ce@3} & place rack & test tube \\
\texttt{e76b2@3} & unscrew & cap & \texttt{155cb@0} & shake & flask & \texttt{7e7aa@0} & rearrange & burner \\
\texttt{e76b2@8} & open gate & microwave & \texttt{81a95@10} & place rack & test tube & \texttt{7e7aa@2} & rearrange & burner \\
\texttt{e76b2@9} & place in & bowl & \texttt{81a95@2} & ignite & ignitor & \texttt{7e7aa@3} & put off & burner cap \\
\texttt{1469a@0} & uncap & porc. cap & \texttt{a5345@1} & shake & test tube & \texttt{783bd@1} & rearrange & fork \\
\texttt{4102c@0} & screw & cap & \texttt{a5345@5} & ignite & ignitor & \texttt{a0687@2} & rearrange & plate \\
\texttt{598a5@0} & cut & knife & \texttt{a5345@6} & remove rack & test tube &  & &  \\
\texttt{67132@0} & cap & porc. cap & \texttt{a5345@8} & heat tube & holder &  & &  \\
\texttt{25281@0} & open laptop & laptop & \texttt{c50ce@0} & hold & beaker &  & &  \\
\texttt{f0408@1} & remove plug & plug & \texttt{cf9b2@4} & rearrange & burner &  & &  \\
\texttt{9cf44@2} & remove usb & usb & \texttt{7e7aa@1} & heat beaker & beaker &  & &  \\
\texttt{9cf44@4} & remove plug & plug & \texttt{9f2cb@1} & shake & test tube &  & &  \\
\texttt{68184@0} & remove usb & usb & \texttt{9f2cb@2} & place rack & test tube &  & &  \\
\texttt{126ad@0} & remove usb & usb & \texttt{c5685@13} & rearrange & flask &  & &  \\
\texttt{155cb@3} & pour & flask & \texttt{c5685@5} & ignite & ignitor &  & &  \\
\texttt{81a95@1} & uncap lamp & burner cap & \texttt{c5685@7} & heat beaker & beaker &  & &  \\
\texttt{81a95@9} & pour & beaker & \texttt{ff220@0} & rearrange & holder &  & &  \\
\texttt{a5345@10} & pour & holder & \texttt{ff220@3} & shake & holder &  & &  \\
\texttt{a5345@9} & pour & beaker & \texttt{ff220@5} & place rack & holder &  & &  \\
\texttt{c50ce@2} & pour & beaker & \texttt{f9335@2} & rearrange & toothbrush &  & &  \\
\texttt{9d98f@1} & stir & rod & \texttt{ff070@0} & rearrange & toothbrush &  & &  \\
\texttt{c5685@0} & unscrew & cap & \texttt{783bd@0} & rearrange & spoon &  & &  \\
\texttt{c5685@1} & pour & bottle &  & &  &  & &  \\
\texttt{c5685@14} & pour & beaker &  & &  &  & &  \\
\texttt{c5685@15} & screw & cap &  & &  &  & &  \\
\texttt{ff220@2} & hold tube & holder &  & &  &  & &  \\
\texttt{f9335@0} & flip open & tp cap &  & &  &  & &  \\
\texttt{f9335@1} & squeeze & tp tube &  & &  &  & &  \\
\texttt{f9335@4} & flip close & tp cap &  & &  &  & &  \\
\texttt{99f9d@0} & open gate & microwave &  & &  &  & &  \\
\texttt{99f9d@1} & place in & mug &  & &  &  & &  \\
\texttt{99f9d@2} & close gate & microwave &  & &  &  & &  \\
\texttt{99f9d@7} & close gate & microwave &  & &  &  & &  \\
\texttt{99f9d@8} & stir & rod &  & &  &  & &  \\
\bottomrule
\end{tabular}
\end{table*}

%% file: IEEEtranBST/references.bib
@article{schulman2017ppo,
  title         = {Proximal Policy Optimization Algorithms},
  author        = {Schulman, John and Wolski, Filip and Dhariwal, Prafulla and Radford, Alec and Klimov, Oleg},
  year          = {2017},
  journal={arXiv},
}

@inproceedings{johannink2018residual,
  title     = {Residual Reinforcement Learning for Robot Control},
  author    = {Johannink, Tobias and Bahl, Shikhar and Nair, Ashvin and Luo, Jianlan and Kumar, Avinash and Loskyll, Matthias and Aparicio Ojea, Juan and Solowjow, Eugen and Levine, Sergey},
  booktitle = {ICRA},
  year      = {2019},
}

@inproceedings{smplx,
  title     = {Expressive Body Capture: {3D} Hands, Face, and Body from a Single Image},
  author    = {Pavlakos, Georgios and Choutas, Vasileios and Ghorbani, Nima and Bolkart, Timo and Osman, Ahmed A. A. and Tzionas, Dimitrios and Black, Michael J.},
  booktitle = {CVPR},
  year      = {2019},
}

@inproceedings{zhou2019continuity,
  title     = {On the Continuity of Rotation Representations in Neural Networks},
  author    = {Zhou, Yi and Barnes, Connelly and Lu, Jingwan and Yang, Jimei and Li, Hao},
  booktitle = {CVPR},
  year      = {2019},
}

@inproceedings{pointnet,
  title     = {{PointNet}: Deep Learning on Point Sets for {3D} Classification and Segmentation},
  author    = {Qi, Charles R. and Su, Hao and Mo, Kaichun and Guibas, Leonidas J.},
  booktitle = {CVPR},
  year      = {2017},
}

@inproceedings{oakinkv2,
  title     = {{OAKINK2}: A Dataset of Bimanual Hands-Object Manipulation in Complex Task Completion},
  author    = {Zhan, Xinyu and Yang, Lixin and Zhao, Yifei and Mao, Kangrui and Xu, Hanlin and Lin, Zenan and Li, Kailin and Lu, Cewu},
  booktitle = {CVPR},
  year      = {2024},
}

@inproceedings{arctic,
  title     = {{ARCTIC}: A Dataset for Dexterous Bimanual Hand-Object Manipulation},
  author    = {Fan, Zicong and Taheri, Omid and Tzionas, Dimitrios and Kocabas, Muhammed and Kaufmann, Manuel and Black, Michael J. and Hilliges, Otmar},
  booktitle = {CVPR},
  year      = {2023},
}

@inproceedings{dexycb,
  title     = {{DexYCB}: A Benchmark for Capturing Hand Grasping of Objects},
  author    = {Chao, Yu-Wei and Yang, Wei and Xiang, Yu and Molchanov, Pavlo and Handa, Ankur and Tremblay, Jonathan and Narang, Yashraj S. and others},
  booktitle = {CVPR},
  year      = {2021},
}

@inproceedings{geort,
  title     = {Geometric Retargeting: A Principled, Ultrafast Neural Hand Retargeting Algorithm},
  author    = {Yin, Zhao-Heng and Wang, Changhao and Pineda, Luis and Bodduluri, Krishna and Wu, Tingfan and Abbeel, Pieter and Mukadam, Mustafa},
  booktitle = {IROS},
  year      = {2025},
}

@inproceedings{anyteleop,
  title     = {{AnyTeleop}: A General Vision-Based Dexterous Robot Arm-Hand Teleoperation System},
  author    = {Qin, Yuzhe and Yang, Wei and Huang, Binghao and Van Wyk, Karl and Su, Hao and Wang, Xiaolong and Chao, Yu-Wei and Fox, Dieter},
  booktitle = {RSS},
  year      = {2023},
}

@inproceedings{dexpilot,
  title     = {{DexPilot}: Vision-Based Teleoperation of Dexterous Robotic Hand-Arm System},
  author    = {Handa, Ankur and Van Wyk, Karl and Yang, Wei and Liang, Jacky and Chao, Yu-Wei and Wan, Qian and Birchfield, Stan and Ratliff, Nathan D. and Fox, Dieter},
  booktitle = {ICRA},
  year      = {2020},
}

@inproceedings{artigrasp,
  title     = {{ArtiGrasp}: Physically Plausible Synthesis of Bi-Manual Dexterous Grasping and Articulation},
  author    = {Zhang, Hui and Christen, Sammy and Fan, Zicong and Zheng, Luocheng and Hwangbo, Jemin and Song, Jie and Hilliges, Otmar},
  booktitle = {3DV},
  year      = {2024},
}

@inproceedings{maniptrans,
  title     = {{ManipTrans}: Efficient Dexterous Bimanual Manipulation Transfer via Residual Learning},
  author    = {Li, Kailin and Li, Puhao and Liu, Tengyu and Li, Yuyang and Huang, Siyuan},
  booktitle = {CVPR},
  year      = {2025},
}

@inproceedings{dexmachina,
  title         = {{DexMachina}: Functional Retargeting for Bimanual Dexterous Manipulation},
  author        = {Zhao, Mandi and Hou, Yifan and Fox, Dieter and Narang, Yashraj and Mandlekar, Ajay and Song, Shuran},
  booktitle     = {ICML},
  year          = {2026},
}

@inproceedings{openvla,
  title     = {{OpenVLA}: An Open-Source Vision-Language-Action Model},
  author    = {Kim, Moo Jin and Pertsch, Karl and Karamcheti, Siddharth and Xiao, Ted and Balakrishna, Ashwin and Nair, Suraj and others},
  booktitle = {CoRL},
  year      = {2024}
}

@inproceedings{dexcap,
  title     = {{DexCap}: Scalable and Portable Mocap Data Collection System for Dexterous Manipulation},
  author    = {Wang, Chen and Shi, Haochen and Wang, Weizhuo and Zhang, Ruohan and Fei-Fei, Li and Liu, C. Karen},
  booktitle = {RSS},
  year      = {2024},
}

@misc{allegrohand,
  author       = {{Wonik Robotics}},
  title        = {Allegro Hand},
  howpublished = {\url{https://www.wonikrobotics.com/research-robot-hand}},
  note         = {Accessed: 2026-06-26}
}

@misc{inspirehand,
  author       = {{Inspire-Robots}},
  title        = {Dexterous Hand ({RH56} Series)},
  howpublished = {\url{https://inspire-robots.store/}},
  note         = {Accessed: 2026-06-26}
}

@inproceedings{spider,
  title = {{SPIDER}: Scalable Physics-Informed Dexterous Retargeting},
  author = {Pan, Chaoyi and Wang, Changhao and Qi, Haozhi and Liu, Zixi and Bharadhwaj, Homanga and Sharma, Akash and Wu, Tingfan and Shi, Guanya and Malik, Jitendra and Hogan, Fran{\c{c}}ois Robert},
    booktitle = {IROS},
  year = {2026}
}

@inproceedings{omniretarget,
  title = {{OmniRetarget}: Interaction-Preserving Data Generation for Humanoid Whole-Body Loco-Manipulation and Scene Interaction},
  author = {Yang, Lujie and Huang, Xiaoyu and Wu, Zhen and Kanazawa, Angjoo and Abbeel, Pieter and Sferrazza, Carmelo and Liu, C. Karen and Duan, Rocky and Shi, Guanya},
  booktitle = {ICRA},
  year = {2026}
}

@article{toporetarget,
  title = {{TopoRetarget}: Interaction-Preserving Retargeting for Dexterous Manipulation},
  author = {Wu, Jielin and Yao, Shenzhe and He, Guanqi and Liu, Xiaohan and Zeng, Zhaoqing and Jiang, Xiangrui and Yang, Han and Zhang, Wentao and Zhao, Hang},
  journal = {arXiv},
  year = {2026},
  archivePrefix = {arXiv},
  eprint = {2606.16272}
}

@article{c2dex,
  title = {{C2Dex}: Contact-Consistent Reconstruction and Retargeting for Dexterous Manipulation from Monocular Video},
  author = {Ren, Jie and Jiang, Zhehao and Yang, Yinhong and Jia, Haorui and Jiang, Han and others},
  journal = {arXiv},
  year = {2026},
  archivePrefix = {arXiv},
  eprint = {2608.07045}
}

@article{regrind,
  title = {A Minimalist Retargeting-Guided Reinforcement Learning Recipe for Dexterous Manipulation},
  author = {Feng, Yunhai and Leung, Natalie and Wang, Jiaxuan and Yang, Lujie and Qi, Haozhi and Culbertson, Preston},
  journal = {arXiv},
  year = {2026},
  archivePrefix = {arXiv},
  eprint = {2607.11874}
}

@InProceedings{r3m,
  title = {{R3M}: A Universal Visual Representation for Robot Manipulation},
  author = {Nair, Suraj and Rajeswaran, Aravind and Kumar, Vikash and Finn, Chelsea and Gupta, Abhinav},
  booktitle = {CoRL},
  year = {2023},
}

@article{egovla,
  title = {{EgoVLA}: Learning Vision-Language-Action Models from Egocentric Human Videos},
  author = {Yang, Ruihan and Yu, Qinxi and Wu, Yecheng and Yan, Rui and Li, Borui and Cheng, An-Chieh and Zou, Xueyan and Fang, Yunhao and others},
  journal = {arXiv},
  year = {2025},
  archivePrefix = {arXiv},
  eprint = {2507.12440}
}

@inproceedings{mimicplay,
title={{MimicPlay}: Long-Horizon Imitation Learning by Watching Human Play},
author={Chen Wang and Linxi Fan and Jiankai Sun and Ruohan Zhang and Li Fei-Fei and Danfei Xu and Yuke Zhu and Anima Anandkumar},
booktitle={CoRL},
year={2023},
}

@inproceedings{egobridge,
  title = {{EgoBridge}: Domain Adaptation for Generalizable Imitation from Egocentric Human Data},
  author = {Punamiya, Ryan and Patel, Dhruv and Aphiwetsa, Patcharapong and Kuppili, Pranav and Zhu, Lawrence Y. and Kareer, Simar and Hoffman, Judy and Xu, Danfei},
  booktitle = {NeurIPS},
  year = {2025}
}

@inproceedings{hop,
  title = {Hand-Object Interaction Pretraining from Videos},
  author = {Singh, Himanshu Gaurav and Loquercio, Antonio and Sferrazza, Carmelo and Wu, Jane and Qi, Haozhi and Abbeel, Pieter and Malik, Jitendra},
  booktitle = {ICRA},
  year = {2025},
}

@article{videomanip,
  title = {Dexterous Manipulation Policies from {RGB} Human Videos via 4{D} Hand-Object Trajectory Reconstruction},
  author = {Chen, Hongyi and Dong, Tony and Wu, Tiancheng and Wang, Liquan and Jangir, Yash and Niu, Yaru and others},
  journal = {arXiv},
  year = {2026},
}

@inproceedings{dextrack,
  title = {{DexTrack}: Towards Generalizable Neural Tracking Control for Dexterous Manipulation from Human References},
  author = {Liu, Xueyi and Adalibieke, Jianibieke and Han, Qianwei and Qin, Yuzhe and Yi, Li},
  booktitle = {ICLR},
  year = {2025}
}

@inproceedings{bidexhd,
  title = {Learning Diverse Bimanual Dexterous Manipulation Skills from Human Demonstrations},
  author = {Zhou, Bohan and Yuan, Haoqi and Fu, Yuhui and Lu, Zongqing},
  booktitle = {AAAI},
  year = {2026}
}

@inproceedings{graspd,
  title = {{Grasp'D}: Differentiable Contact-Rich Grasp Synthesis for Multi-Fingered Hands},
  author = {Turpin, Dylan and Wang, Liquan and Heiden, Eric and Chen, Yun-Chun and Macklin, Miles and Tsogkas, Stavros and Dickinson, Sven J. and Garg, Animesh},
  booktitle = {ECCV},
  year = {2022},
}

@inproceedings{contactopt,
    author={Grady, Patrick and Tang, Chengcheng and Twigg, Christopher D. and Vo, Minh and Brahmbhatt, Samarth and Kemp, Charles C.},
    title = {{ContactOpt}: Optimizing Contact to Improve Grasps},
    booktitle = {CVPR},
    year = {2021}
}

@article{pyslsqp,
  title = {{PySLSQP}: A transparent Python package for the {SLSQP} optimization algorithm modernized with utilities for visualization and post-processing},
  author = {Anugrah Jo Joshy and John T. Hwang},
  journal = {Journal of Open Source Software},
  publisher = {The Open Journal},
  year = {2024},
  url = {https://doi.org/10.21105/joss.07246},
  doi = {10.21105/joss.07246},
}

@book{numericaloptim,
author = {Nocedal, Jorge. and Wright, Stephen J.},
booktitle = {Numerical optimization},
isbn = {0387303030},
publisher = {Springer},
title = {Numerical optimization},
year = {2006},
}

@article{deepmimic,
  title   = {{DeepMimic}: Example-Guided Deep Reinforcement Learning of Physics-Based Character Skills},
  author  = {Peng, Xue Bin and Abbeel, Pieter and Levine, Sergey and van de Panne, Michiel},
  journal = {TOG},
  year    = {2018},
}

@article{resmimic,
  title   = {{ResMimic}: From General Motion Tracking to Humanoid Whole-Body Loco-Manipulation via Residual Learning},
  author  = {Zhao, Siheng and Ze, Yanjie and Wang, Yue and Liu, C. Karen and Abbeel, Pieter and Shi, Guanya and Duan, Rocky},
  journal = {arXiv},
  year    = {2025},
}

@article{beyondmimic,
  title={{BeyondMimic}: From Motion Tracking to Versatile Humanoid Control via Guided Diffusion},
  author={Liao, Qiayuan and Truong, Takara E. and Huang, Xiaoyu and Gao, Yuman and Tevet, Guy and Sreenath, Koushil and Liu, C. Karen},
  journal={Science Robotics},
  year={2026},
}

@article{pi07,
  title  = {$\pi_{0.7}$: a Steerable Generalist Robotic Foundation Model with Emergent Capabilities},
  author = {{Physical Intelligence}},
  journal   = {arXiv},
  year   = {2026}
}
